\documentclass[runningheads]{llncs}
\usepackage[T1]{fontenc}
\usepackage{graphicx}
\usepackage{multirow}
\usepackage{multicol}
\usepackage{amssymb}
\usepackage{diagbox}
\usepackage{subfigure}
\usepackage{colortbl}
\usepackage[table]{xcolor}
\begin{document}
\title{Versatile Defense Against Adversarial Attacks on Image Recognition}
%
%\titlerunning{Abbreviated paper title}
% If the paper title is too long for the running head, you can set
% an abbreviated paper title here
%
\author{Haibo Zhang\inst{1} \and
Zhihua Yao\inst{2} \and
Kouichi Sakurai\inst{3}}
\authorrunning{H. Zhang et al.}
% First names are abbreviated in the running head.
% If there are more than two authors, 'et al.' is used.
%

\institute{Department of Information Science and Technology, Graduate School of Information Science and Electrical Engineering, Kyushu University, Japan \\
\email{zhang.haibo.892@s.kyushu-u.ac.jp}\and
Faculty of Economics and Business Administration, The University of Kitakyushu
\email{zhihuayao@alumni.usc.edu} \and
Department of Information Science and Technology, Faculty of Information Science and Electrical Engineering, Kyushu University, Japan \\
\email{sakurai@inf.kyushu-u.ac.jp}}
\maketitle              % typeset the header of the contribution
\begin{abstract}
Adversarial attacks present a significant security risk to image recognition tasks. Defending against these attacks in a real-life setting can be compared to the way antivirus software works, with a key consideration being how well the defense can adapt to new and evolving attacks. Another important factor is the resources involved in terms of time and cost for training defense models and updating the model database. Training many models that are specific to each type of attack can be time-consuming and expensive. Ideally, we should be able to train one single model that can handle a wide range of attacks. It appears that a defense method based on image-to-image translation may be capable of this. The proposed versatile defense approach in this paper only requires training one model to effectively resist various unknown adversarial attacks. The trained model has successfully improved the classification accuracy from nearly zero to an average of 86\%, performing better than other defense methods proposed in prior studies. When facing the PGD attack and the MI-FGSM attack, versatile defense model even outperforms the attack-specific models trained based on these two attacks. The robustness check also shows that our versatile defense model performs stably regardless with the attack strength.

\keywords{Generative adversarial network \and Image-to-image translation  \and Adversarial attack \and Versatile defense.}
\end{abstract}
\section{Introduction}

Convolutional Neural Networks (CNNs) represent a pivotal class of deep learning models and have seen widespread application in diverse image recognition tasks, ranging from object detection to facial recognition and autonomous driving. Nevertheless, these models harbor a significant susceptibility to adversarial attacks as evidenced by seminal studies, such as the groundbreaking research conducted by Goodfellow et al.\cite{goodfellow2014explaining}. These adversarial attacks leverage meticulously crafted, subtle alterations to an image, which, although virtually imperceptible to the human eye, can confound the model into committing misclassification errors.

\begin{figure}
\includegraphics[width=\textwidth]{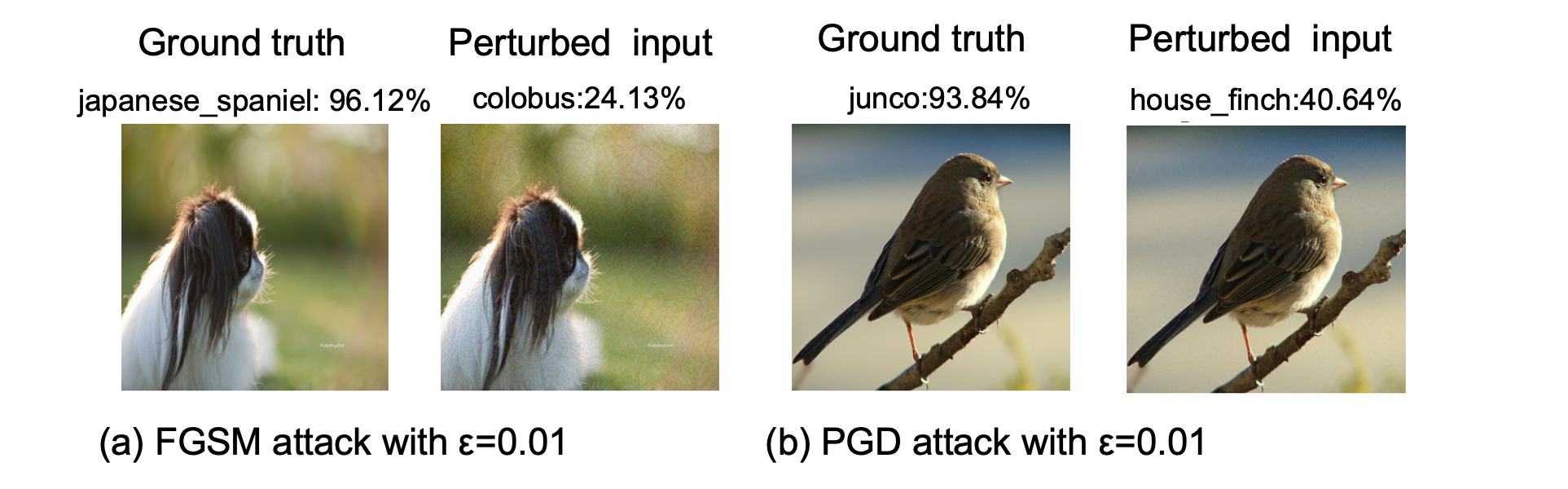}
\caption{Adversarial examples of FGSM attack and PGD attack. In part (a), the original image is correctly classified as a Japanese\_spaniel with a confidence of 96.12\%, but the perturbed image, crafted by the FGSM attack, is misclassified as a colobus with a confidence of 24.13\%. In part (b), the original image is correctly classified as a junco with a confidence of 93.84, but the perturbed image, crafted by the PGD attack, is misclassified as a house\_finch with a confidence of 40.64\%.} \label{fig1}
\end{figure}

The vulnerabilities of machine learning models have serious consequences, especially in important applications like facial recognition and autonomous driving. This makes it crucial to design and put in place strong defenses against adversarial attacks. This is particularly true for machine learning systems used in safety-critical situations, a point also emphasized by Xu et al.\cite{xu2020adversarial}. Strengthening these models against such attacks is a key step for the safe and widespread use of these machine learning techniques. 

Defending against adversarial attacks can be likened to the operation of antivirus software, and in the assessment of such defensive methods, generalizability is a paramount consideration given the continuous evolution and proliferation of novel adversarial attacks. Another pivotal factor to consider is the resource expenditure in terms of training cost and processing time. Training numerous attack-specific models and updating the model database is a burdensome process that incurs significant time and monetary costs. An ideal defense method would be to train a single versatile model exhibiting robust generalizability against unforeseen attacks. It is found that a defense method based on image-to-image translation holds the potential to achieve this objective. We will demonstrate this in the rest of the paper.

\subsection{Related Works and Background}
The method discussed in this paper is a defense mechanism against image adversarial attacks based on image-to-image translation scheme. Hence, in this section, we will comprehensively discuss the background information and related works of image-to-image translation technology, adversarial attacks, and some major defense methods.

\subsubsection{Adversarial Attacks}
Adversarial attacks involve subtly modifying the input data to a model, which might be almost imperceptible to human eyes, but could lead the model to misclassify the image. Some representative attacks dealt with in this paper including Fast Gradient Sign Method (FGSM)\cite{goodfellow2014explaining}, Basic Iterative Method (BIM)\cite{kurakin2018adversarial}, Projected Gradient Descent (PGD)\cite{madry2017towards}, Carlini \& Wagner (C\&W)\cite{carlini2017towards}, Momentum Iterative Fast Gradient Sign Method (MI-FGSM)\cite{dong2018boosting}, DeepFool\cite{moosavi2016deepfool}, and AutoAttack\cite{croce2020reliable}.

\subsubsection{Defense Mechanisms}
To mitigate the impact of adversarial attacks, more and more researchers have proposed various defense mechanisms. These methods are primarily divided into two categories: reactive defenses and proactive defenses. Reactive defenses aim to detect and discard adversarial examples, while proactive defenses strive to train the model to resist adversarial perturbations. For instance, adversarial training, a form of proactive defense, has shown promise. It involves augmenting the training data with adversarial examples and has been shown to significantly enhance the model's robustness against adversarial attacks \cite{goodfellow2014explaining,madry2017towards}.

Despite these advancements, finding universally applicable, efficient, and reliable defense strategies remains an open challenge, as evidenced by recent works showing that many existing defense mechanisms can be circumvented by properly designed adversarial attacks \cite{carlini2017adversarial}, \cite{athalye2018obfuscated}.

\subsubsection{Image-to-image Translation}
The image-to-image translation is a subfield of computer vision, primarily focusing on converting images from one domain to another using method like Generative Adversarial Networks (GANs)\cite{goodfellow2020generative}. Image-to-image translation can be viewed as a variant of Conditional Generative Adversarial Networks (cGANs)\cite{mirza2014conditional}. Among these, Pix2pix\cite{isola2017image} and StyleGAN\cite{karras2019style} algorithms are considered representative image translation methods. They have been successfully applied in various tasks, such as transforming satellite images into maps, converting daytime images into nighttime images, etc .\cite {kurakin2018adversarial}.

The image reconstruction approach implemented in this study applies the Pix2pix algorithm\cite{isola2017image} as the basic method, which leverages the power of cGANs. The Pix2Pix model consists of two main components: a generator and a discriminator. The generator, constructed with a U-Net-like architecture, aims to create images that look real, while the discriminator, a PatchGAN classifier, has the job of differentiating between real and generated images.

\subsection{Our Contributions}

%In this study, we conduct comprehensive experiments and assessments on an innovative image reconstruction method. This method, anchored in the realm of image-to-image translation technology, aiming to serve as a universal defense strategy against perturbation-based adversarial attacks. In the face of these increasingly sophisticated attacks, this paper introduces a universally applicable defensive methodology that necessitates the training of only a single model to counteract various potentially unknown, adversarial attacks. The technique fundamentally restores the adversarially tampered images back to their original, undisturbed state, thereby enhancing the model's capability to classify them accurately.

The major contributions of this study are as follows:
\begin{itemize}
    \item We prove that image-to-image translation based defense method has better generalizability to unkonwn adversarial attacks than other existing defense methods. 
    \item We show that using adversarial samples generated by multiple attacks in the training process outperforms models trained using one specific attack.
    \item The versatile defense model we trained successfully recovered the classification accuracy from nearly 0 to an anverage of 86\%, which is better than any other defense methods proposed by previous studies. When against the PGD attack and MI-FGSM attack, the performance is even better than attack-specific models trained by these two attacks.
    \item As a robustness check, we evaluate the model in different attack strength, the result show that the versatile defense model is performs stably regardless of the attack strength. We also provide a detailed evaluation of the experimental results using MAE (Mean Absolute Error) values, PSNR (Peak Signal-to-Noise Ratio) values.  .
 
\end{itemize}

\section{Our Proposal}
In the field of deep learning, adversarial attacks have become a serious issue, as they can easily deceive well-trained neural networks. To solve this issue, many methods\cite{samangouei2018defense,mustafa2019image,prakash2018deflecting,zhang2021conditional} have been proven effective against adversarial attacks. However, most previous studies trained their model on one specific attack and aimed to defend that specific attack.  Consequently, a natural question arises: Can we train a single, versatile model that is capable of resisting multiple adversarial attacks? The answer is "YES". In this section, we illustrate our attempts to train a versatile model.

\subsection{Research Motivation}
In previous studies, a unique model $M_i$ was trained to defend against each type of adversarial attack $i$ ($i \in {1, 2, ..., n}$). Therefore, a total of $n$ models are required to resist all types of adversarial attacks, which can be mathematically expressed as:

$$ M = \{M_1, M_2, ..., M_n\}$$

However, our goal is to train a single model, $M'$, that can withstand all $n$ types of adversarial attacks. Therefore, we hope that, although there is only one model, its function is equivalent to the set of models $M$:

$$ M' = M $$

This implies that our objective is to verify whether $M'$ meets our expectations, i.e., whether $M'$ achieves a satisfactory performance when encountering multiple adversarial attacks, or even better performance when compared with models that are trained for one specific attack only. 
%If $M'$ can successfully cope with all types of adversarial attacks, then it would mean that our hypothesis is correct — a single model is sufficient to resist multiple adversarial attacks.

\subsection{Model Description}

\begin{figure}
\includegraphics[width=\textwidth]{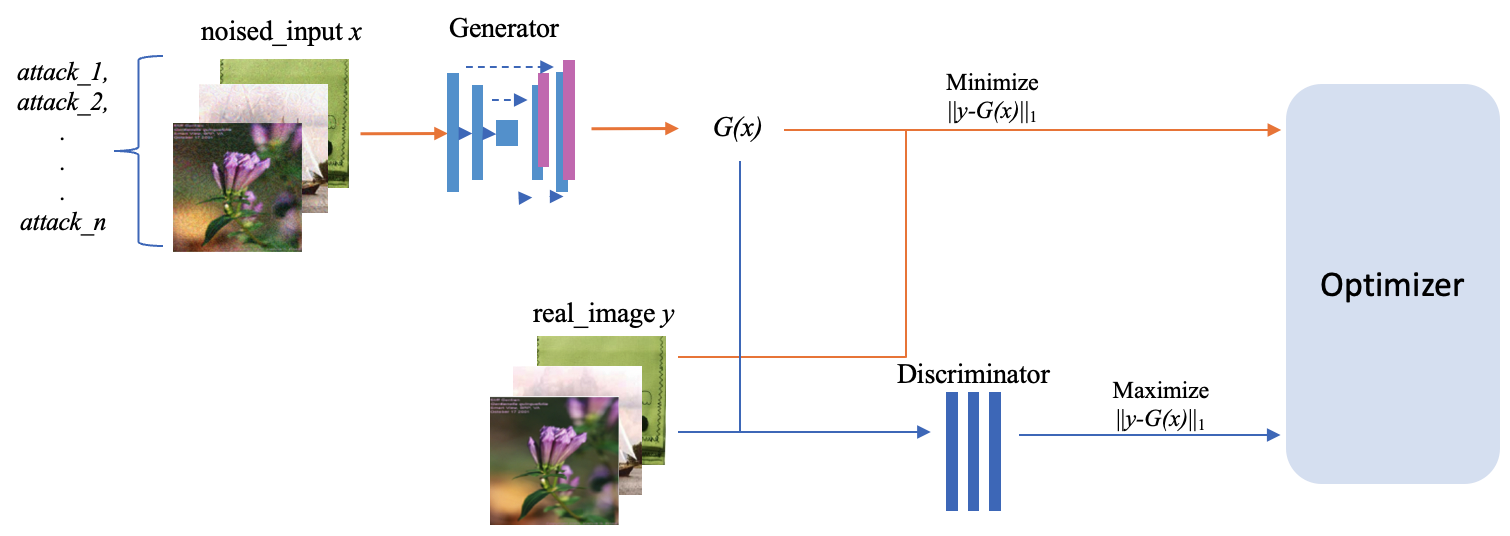}
\caption{The training progress of image reconstruction method.} \label{fig2}
\end{figure}

Our model is based on the algorithm proposed in the study\cite{zhang2021conditional}, which adds perceptual loss to the basic Pix2pix algorithm and use it to reconstruct perturbed images. We use the same image reconstruction method, as Fig.2 shows, but during the training process, we utilize adversarial samples that are generated by multiple adversarial attacks.

The algorithm is trained by optimizing the following objective function, which is a combination of the adversarial loss (for the GAN), an $L1$ loss (also known as pixel loss, which encourages the generated image to be structurally similar to the target image) and perceptual loss (which measures the difference between two images based on their perceptual similarity):

\begin{equation}
    \mathcal{L}_{cGAN}(G,D) = \mathbb{E}_{x,y}[\log D(x,y)] + \mathbb{E}_{x,z}[\log (1-D(x, G(x,z)))]
\end{equation}

\begin{equation}
    \mathcal{L}_{L1}(G) = \mathbb{E}_{x,y,z}[||y-G(x,z)||_1]
\end{equation}

\begin{equation}
    {\cal{L}}_{perceptual} (G) = \sum_k a_k || V_k (y) - V_k (G(x, z))||_1
\end{equation}

\begin{equation}
    G^* = \arg\min_G \max_D \mathcal{L}_{cGAN}(G,D) + \lambda{1} \mathcal{L}_{L1}(G) + \lambda{2}  {\cal{L}}_{perceptual} (G)
\end{equation}

In these equations, $x$ is the input image, $y$ is the target image, $z$ is a random noise vector, $G$ is the generator, $D$ is the discriminator, $||.||_1$ denotes the $L1$ norm (which measures the absolute differences between the target and the generated images). For the perceptual loss, $V$ denotes the pre-trained model VGG19\cite{simonyan2014very}, $k$ is the kth layer in both generated image $G(x, z)$ and target image $y$, $ak$ refers to taking the mean of the differences between the features extracted from the target image and the generated image. The $\lambda{1}$ and $\lambda{2}$ are weights of the $L1$ loss and perceptual loss relative to the adversarial loss. The objective of the training is to find the generator $G$ that minimizes this combined loss.

%To assess the universality of the model, we evaluate it from two aspects: transferability and generalization ability. Transferability refers to whether the model, when trained against one attack, can effectively counter other attacks; generalization ability, on the other hand, refers to the model's capability to handle unseen attack samples. Through these two indicators, we aim to comprehensively evaluate the model's universality.

\section{Experiments}

\subsection{Training Preparation}

\subsubsection{Implemetation details}
The following experiment was conducted using an NVIDIA GeForce RTX 3060 Ti. The chosen programming language is Python, and the utilized deep learning framework is tensorflow\_2.
\subsubsection{Target model}
We choose the pre-trained Inception-v3\cite{szegedy2016rethinking} model (on the ImageNet dataset) as the target model in our study and implement adversarial attacks on it.
\subsubsection{Dataset}
Regarding the training dataset, we randomly selected 18 images from each of the 1000 categories in the ILSVRC2012 ImageNet\cite{ILSVRC15} training set. Every set of 3 images underwent one adversarial attack (a total of 6 types of attacks), resulting in a training set encompassing 18,000 images. As for the test set, we randomly chose 5 images from each of the 1000 categories within ImageNet's validation set, totaling 5,000 images. During each testing session, these 5,000 images serve as the benchmark for conducting attack-recovery tests. We resized all images to 256*256 to comply with the input size of our model.
\subsubsection{Attack models}
We implemented six common attacks, which are FGSM, BIM, PGD, C\&W, MI-FGSM and AutoAttack, using the attack function provided by Cleverhans\cite{papernot2018cleverhans}. The AutoAttack was executed using the attack function from Adversarial Robustness Toolbox v1.2.0 (ART)\cite{art2018}. Our experimental findings suggested that the quality of the trained generative model improved with increasing attack degree within the training dataset. Therefore, we raised the $\epsilon$ to 16/255 for FGSM, BIM, PGD, MI-FGSM and AutoAttack in an effort to enhance the image generation capabilities of the generative model. We set the norm value to np.inf, indicating an adversarial attack governed by the $L\infty$-norm constraint. Additionally, we conducted 40 iterations in total (nb\_iter=40), with each iteration updating the perturbation in the gradient direction of the loss function at a step size (stride) of 0.01 (eps\_iter=0.01). As for the C\&W attack, we adopt $L2$-norm constraint as recommended in the original paper.

\subsection{Multi-steps Training}
In this study, we utilize the multi-step training method for training the image reconstruction model. Traditional neural network models establish pre-defined hyperparameters and assign fixed weights to the loss functions. In contrast, our investigation involves two loss functions, pixel loss and perceptual loss, requiring the tuning of their associated weights, denoted as $\lambda 1$ and $\lambda 2$, respectively. It is difficult to determining the best combination of $\lambda 1$ and $\lambda 2$ at the same time, so adopt a multi-step training process to adjust these values sequentially. 

The rationale behind this method is to first ensure that the reconstructed images capture the high-level structural and semantic content of the original images (which is the strength of perceptual loss), and then fine-tune the details at the pixel level (which is the strength of pixel loss).

Specifically, we establish a total training epoch count of 100, during which the weights are adjusted as 1) from epoch 0 to 39, both $\lambda 1$ and $\lambda 2$ are set to 100, enhancing the impact of the perceptual loss during this training phase, allowing the model to focus more on preserving the structural and semantic content of the images; 2) from epoch 40 to 69, while  $\lambda 1$ remains at 100,  $\lambda 2$ is reduced to 50, gradually diminishing the influence of perceptual loss; 3) from epoch 70 to 99,  $\lambda 1$ is kept at 100 and  $\lambda 2$ is further decreased to 1, shifting the focus to the pixel loss, helping refine the details, enabling the model to produce images that are numerically close to the target images.

Empirical evidence demonstrates that compared to a fixed weight setting throughout the training, such as  $\lambda 1$ equals 100 and  $\lambda 2$ equals 1, our multi-step training approach yields more preferable results in image reconstruction.

\subsection{Evaluations}

\subsubsection{Model Generalizability}
Before initiating the training of our versatile defense model, $M'$, we first conducted an evaluation of the model's generalizability across multiple transfer tasks, which involves various types of adversarial attacks. The term 'generalizability' is an important concept in machine learning, which refers to the ability of a model to apply what it has learned from its training data to new, unseen data.  Specifically, a model with strong generalizability can not only perform excellently on the training set, but also maintain high performance on data it has never encountered. The generalizability of a model is a crucial measure of model quality as it provides insights into whether the model can adapt to various data distribution changes in real-world scenarios.

In our experiment, we utilized six pre-trained image reconstruction models, namely $M_{FGSM}$, $M_{BIM}$, $M_{PGD}$, $M_{MI}$, $M_{C\&W}$, and $M_{AA}$, to undertake image restoration tasks under seven different adversarial attacks (where DeepFool is designated as an unknown attack outside of the training set). Our goal was to assess whether these models could maintain their good recovery performance when faced with different adversarial attacks than which they are trained on.

\begin{table}
\renewcommand\arraystretch{1.3}
\tabcolsep=0.17cm
\caption{The classification accuracy of six pre-trained image reconstruction models, namely $M_{FGSM}$, $M_{BIM}$, $M_{PGD}$, $M_{MI}$, $M_{C\&W}$, and $M_{AA}$, was evaluated against their corresponding adversarial attacks, along with an unknown attack called DeepFool. The best restoration performance for each attack is highlighted in bold.}\label{tab1}
\begin{center}
\scriptsize
\begin{tabular}{c|c|c|c|c|c|c|c}
\hline
Attacks($\epsilon=0.01$) & No defense & $M_{FGSM}$ & $M_{BIM}$ & $M_{PGD}$ & $M_{MI}$ & $M_{C\&W}$ & $M_{AA}$\\
\hline
Clean & 100\% & 95.3\% & \textbf{96.5\%} & 96.3\% & 95.6\% & 93.1\% & 91.7\% \\
FGSM & 25.8\% & \textbf{83.2\%} & 70.6\% & 80.7\% & 85.74\% & 60.35\% & 80.6\% \\
BIM & 1.8\% & 75.72\% & \textbf{90.42\%} & 73.2\%& 71.3\% & 75.8\% & 70.3\% \\
PGD & 2.4\% & 84.3\% & 83.7\% & \textbf{88.4\%} & 77.45\% & 76.6\% & 83.14\% \\
MI-FGSM & 0.1\% & 83.6\% & 77.36\% & 65.3\% & \textbf{86.34\%} & 64.5\% & 67.8\% \\
C\&W & 0\% & 77.1\% & 77.4\% & 78.5\% & 76.4\% & \textbf{89.1\%} & 78.1\% \\
AutoAttack & 0.5\% & 67.1\% & 70.15\% & 77.8\% & 79.5\% & 79.7\% & \textbf{88.4\%}\\
\rowcolor{lightgray!50}
DeepFool & 0.4\% & 78.56\% & 73.3\% & 76.1\% & 79.1\% & 70.3\% & \textbf{80.7\%} \\
\hline
Average & - & 80.61\% & 79.93\% & 79.54\% & \textbf{81.43\%} & 76.18\% & 80.09\% \\
\hline
\end{tabular}
\end{center}
\end{table}

As illustrated in Table 1, all six models improved the accuracy of image classification to some extent when confronted with adversarial attacks outside of their training sets. This implies that these image reconstruction models can not only effectively defend against adversarial attacks they encountered during training, but also demonstrate a certain level of defense capability against new adversarial attacks they have never seen before. This is a manifestation of the model's generalizability.

Our next test is to see whether a model trained on multiple attacks performs better than any models trained on one single attack. If adding attacks in the training process do enhance the model's generalizability, we can train a versatile model that can perform better than any attack-specific defense models when facing a wide range of adversarial attacks. This kind of model is more useful in the real-life scenario.

%\subsubsection{Defense Versatility}
%However, even if a model performs well under some unseen adversarial attacks, it does not necessarily mean it possesses universal defense capability. Thus, even if strong generalizability may hint at a model's defensive performance under unknown adversarial attacks, it does not equate to demonstrating the model's universal defense capability. To prove a model's universal defense capability, an evaluation of the model's performance under a broader, more diverse range of adversarial attacks is required.

\begin{table}
\renewcommand\arraystretch{1.3}
\tabcolsep=0.17cm
\caption{The classification accuracy of our versatile defense method compared with attack-specific defense models and other state-of-the-art defense methods. The abbreviations denote the following defensive methods: 'Random' refers to the Random Resizing Method, 'WD' corresponds to the Wavelet Denoising Method, 'PD' represents the Pixel Deflection Method, and 'SP' signifies the Super-Resolution Method.  They are evaluated against six adversarial attacks, along with an unknown attack called DeepFool. The best restoration performance for each attack is highlighted in bold.}\label{tab2}
\begin{center}
\scriptsize
\begin{tabular}{c|c|c|c|c|c|c}
\hline
Attacks($\epsilon=0.01$) & No defense & Ours & Atk-specific & Random & WD+PD & WD+SR \\
\hline
Clean & 100\% & 92.1\% & 94.75\% & \textbf{97.3\%} & 96.5\% & 86.3\% \\
FGSM & 25.8\% & 80.4\% & \textbf{83.2\%} & 51.3\% & 37.5\% & 70.7\% \\
BIM & 1.8\% & 87.3\% & \textbf{90.42\%} & 19.72\% & 28.9\% & 73.2\% \\
PGD & 2.4\% & \textbf{86.76\%} & 88.4\% & 21.3\% & 29.2\% & 79.4\% \\
MI-FGSM & 0.1\% & \textbf{89.3\%} & 86.34\% & 14.5\% & 15.44\% & 70.58\% \\
C\&W & 0\% & 88.4\% & 89.1\% & \textbf{90.1\%} & 35.4\% & 87.5\% \\
AutoAttack & 0.5\% & 83.7\% & \textbf{88.4\%} & 20.1\% & 15.6\% & 71.8\% \\
\rowcolor{lightgray!50}
DeepFool & 0.4\% & 80.3\% & 76.34 & \textbf{88.9\%} & 37.8\% & 73.36\% \\
\hline
Average & - & 86.03\% & \textbf{87.12\%} & 50.4\% & 37.04\% & 76.6\% \\
\hline
\end{tabular}
\end{center}
\end{table}

In Table 2, we compare the classification accuracy the versatile model we trained with attack-specific defense models that are trained using image-to-image translation\cite{zhang2021conditional} and other representative adversarial attack defense methods. The defense methods for comparison include Random Resizing\cite{xie2017mitigating}, Pixel Deflection with Wavelet Denoising\cite{prakash2018deflecting}, and Super-Resolution with Wavelet Denoising\cite{mustafa2019image}. According to Table 2, our versatile model is able to recover the classification accuracy to an average of 86\%, which is higher than the best performer of attack-specific model (81.43\% when trained on MI-FGSM, according to Table 1). Also, when encountering PGD attack and MI-FGSM attack, the versatile model outperforms the attack-specific models that are trained to defend these two attacks. The difference in average accuracy rate is only 1\%, which is small enough considering we only need to train one model but not seven models. In the real-life setting, versatile model is a more feasible choice than training many attack-specific defense models. As for other defense methods proposed by previous studies, the best performer is WD+SR, which achieves an average accuracy rate of 76.6\%, 10\% lower than our versatile model.

\begin{figure}[htbp]
  \centering
  \subfigure[]{
    \includegraphics[width=0.45\textwidth]{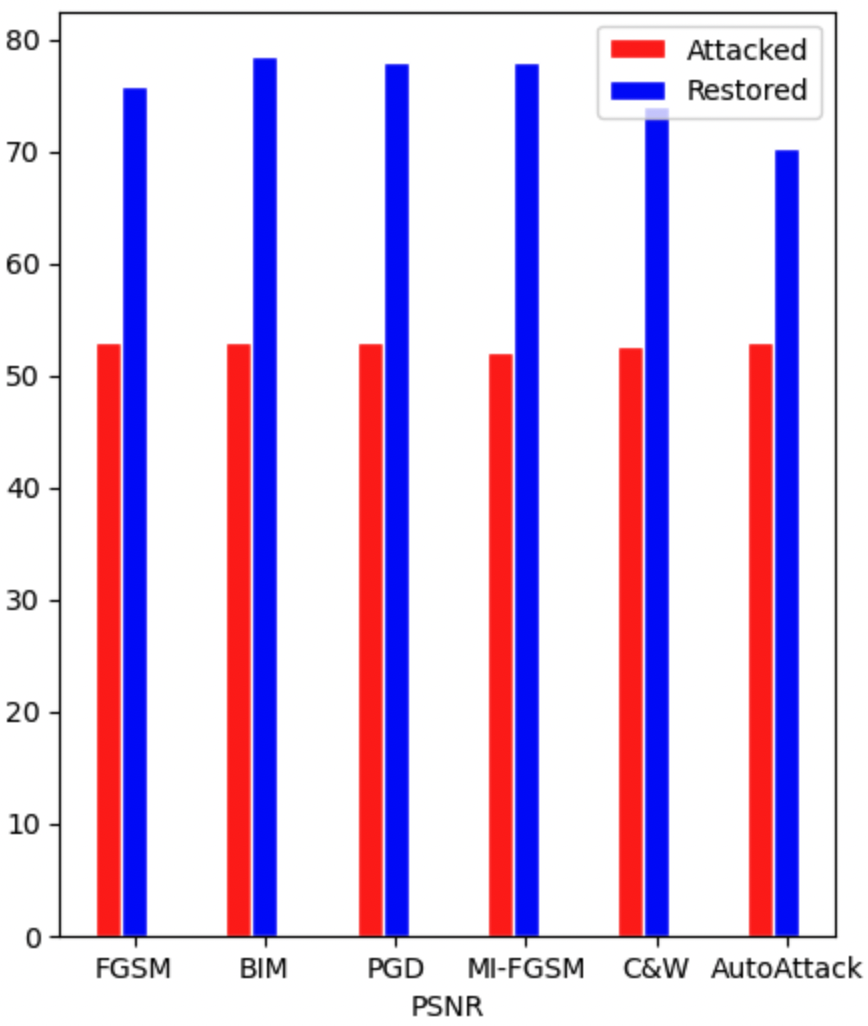}
    \label{fig:subfig1}
  }
  \hfill
  \subfigure[]{
    \includegraphics[width=0.45\textwidth]{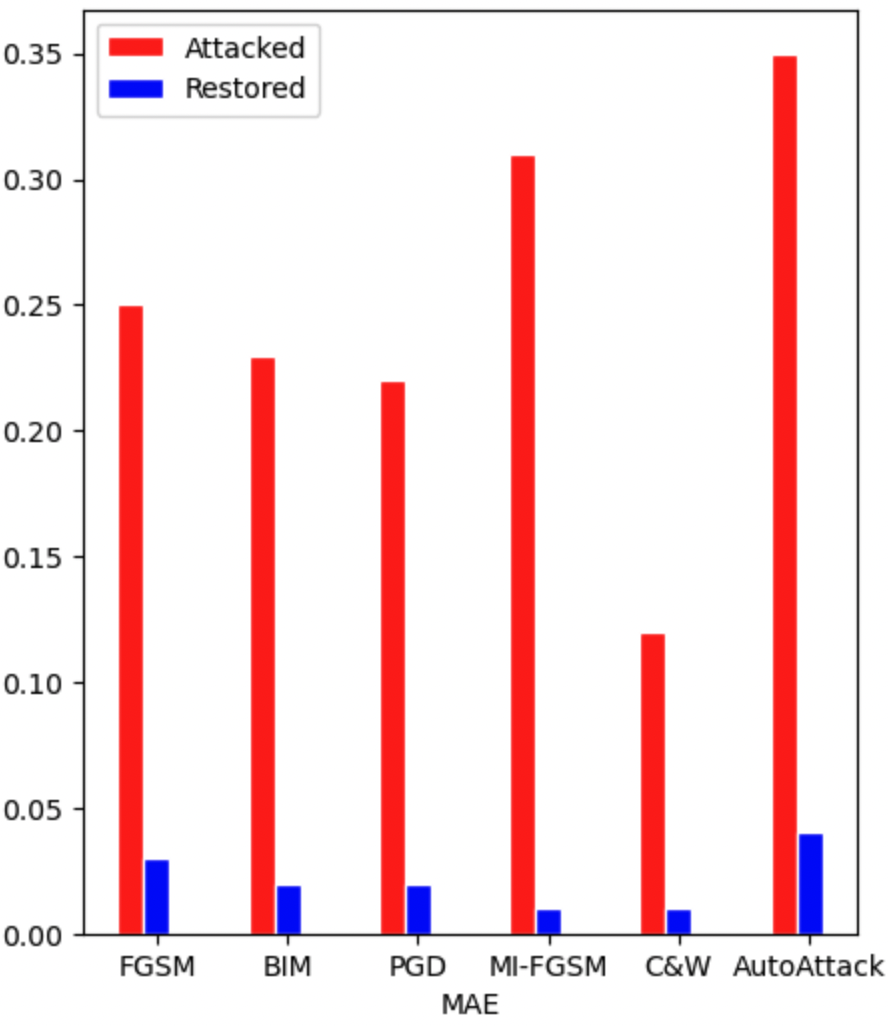}
    \label{fig:subfig2}
  }
  \caption{The computation of the PSNR and MAE values for both the images subjected to six types of adversarial attacks and those reconstructed by the universal defense model when compared to the original images.}
  \label{fig:twosubfigures1}
\end{figure}

\begin{figure}[!ht]
  \centering
  \subfigure[]{
    \includegraphics[width=0.45\textwidth]{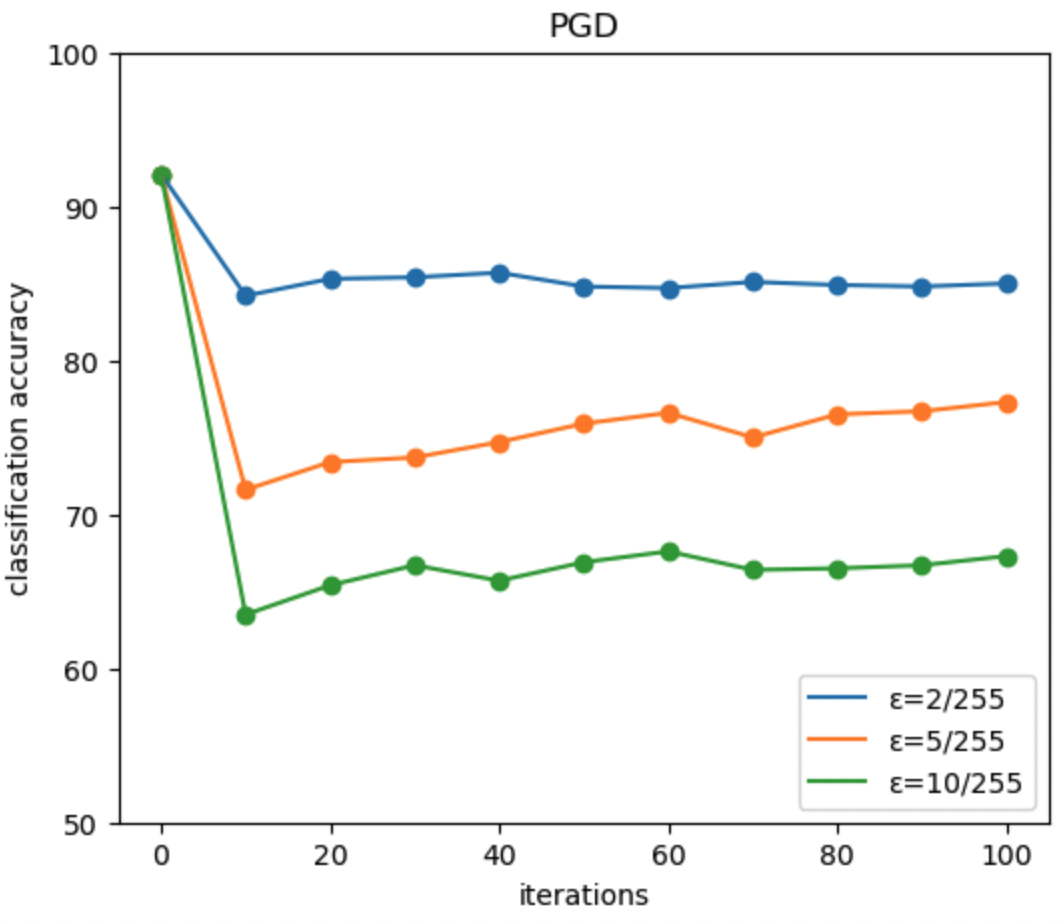}
    \label{fig:sfig1}
  }
  \hfill
  \subfigure[]{
    \includegraphics[width=0.45\textwidth]{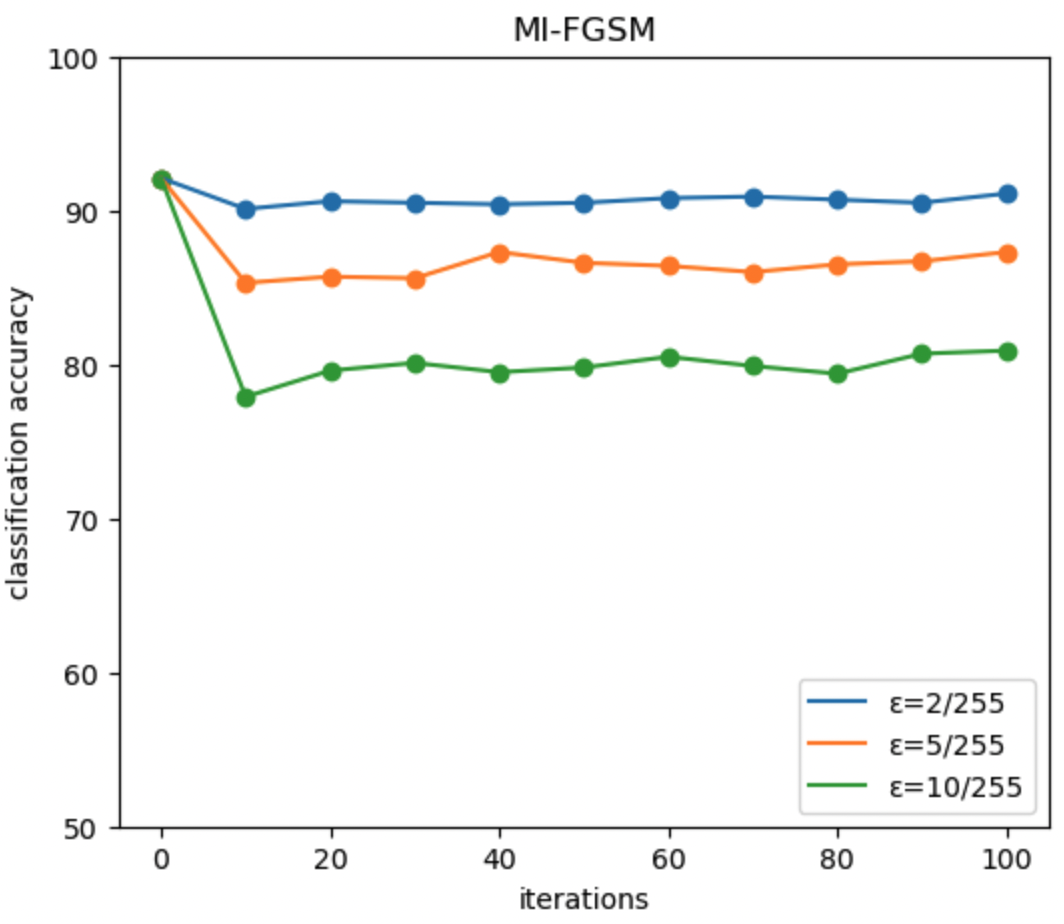}
    \label{fig:sfig2}
  }
  \caption{Robustness Check using the PGD attack and the MI-FGSM attack. To simulate different attack strength, we gradually changing the iteration number from 10 to 100, and the $\epsilon$ includes 2/255, 5/255, and 10/255.}
  \label{fig:twosubfigures2}
\end{figure}

\subsubsection{Quantitative Evaluation}
We also evaluate the performance of our model using two established quantitative metrics: Mean Absolute Error (MAE) and Peak Signal-to-Noise Ratio (PSNR). It is evident from Fig. 3 that our model has been largely successful in reconstructing images after adversarial attacks, thus demonstrating its prominent defensive and recovery capabilities.

\subsubsection{Robustness Check}
As a robustness check, we evaluate the stability of the model performance when facing different attack strengths. We use the PGD attack and the MI-FGSM attack, gradually changing the iteration number from 10 to 100, and the $\epsilon$ includes 2/255, 5/255, and 10/255. As we show in Fig.4, the recovered classification accuracy rate is nearly stable regardless of the iteration numbers.

\begin{figure}
\includegraphics[width=\textwidth]{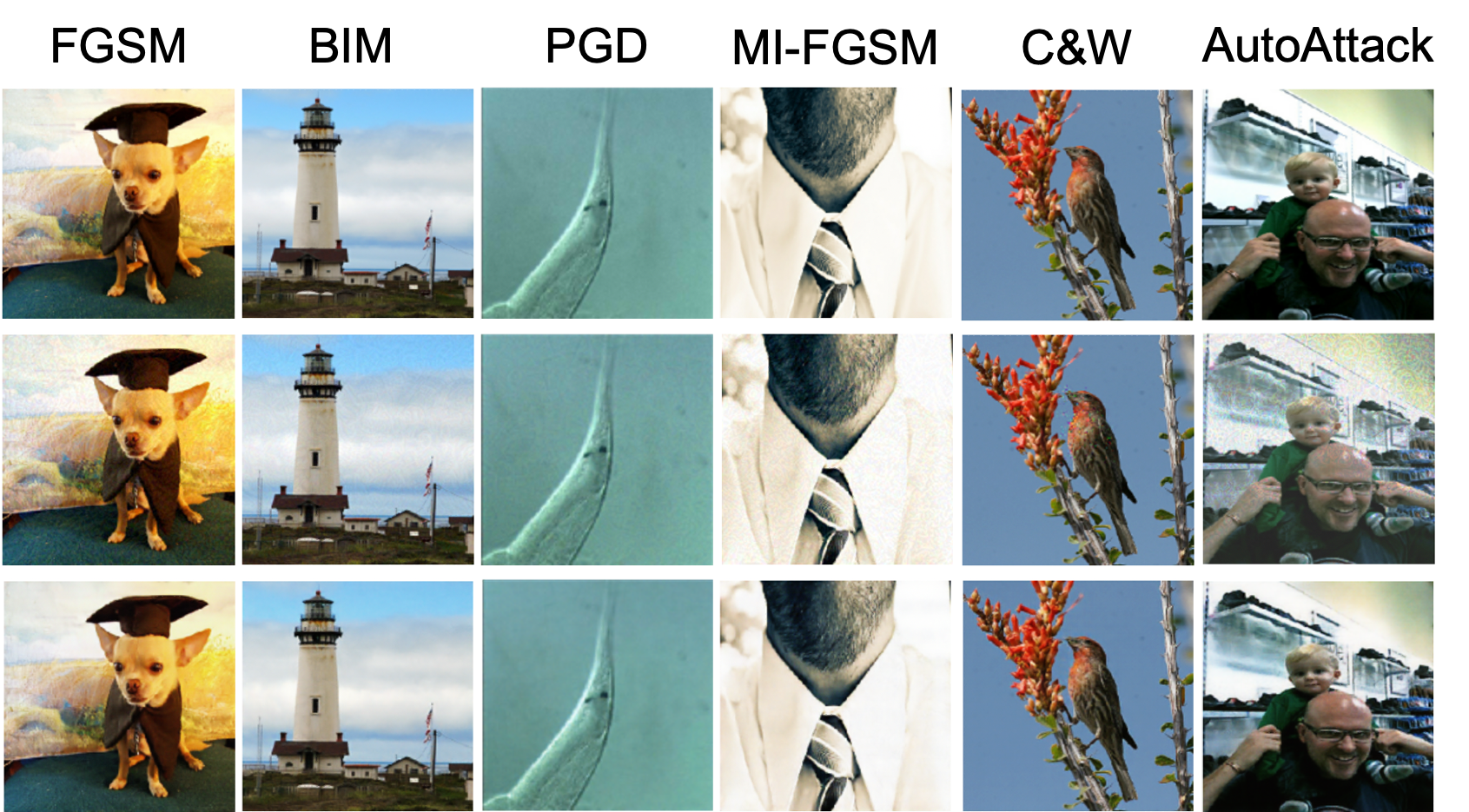}
\caption{ The visual display of images under six different adversarial attack scenarios. The layout is as follows: 1) the first row consists of the original clean images; 2) the second row displays the corresponding images post-attack; and 3) the third row exhibits the images after they have been processed and restored by our universal defense model.} \label{fig3}
\end{figure}

\subsubsection{Visual Effects}
Fig. 5 provides a visual display of images under six different adversarial attack scenarios. This graphic representation allows for an intuitive understanding of the impact of the attacks and the  of our defense model in recovering the original image quality.

\subsection{Supplement Explanation on Experimental Details}
%\subsubsection{Generalizability}
%Generalizability and universal defense capability are two interrelated but not identical concepts. The generalization ability of a model refers to the model's performance when facing unseen data, whereas the universal defense capability refers more specifically to the model's resilience when faced with various types of adversarial attacks. A model with strong generalization ability can still perform well when encountering data not included in the training set. This is because such a model can capture the underlying structures and patterns of the data, rather than merely memorizing specific instances from the training set. Therefore, this model type is more likely to retain good performance when faced with new, unseen adversarial attacks.
\subsubsection{Direct Tensor Data Processing}
In our study, we discovered that the recognition accuracy of image classifiers significantly increased if perturbed images were saved and then reloaded. However, these alterations remained indiscernible to human eyes. Numerous potential causes could underlie this phenomenon, including the inherent instability of adversarial attacks, loss of image information during the saving process (despite our use of lossless PNG format), and loss of information due to normalization during image loading, among other unidentified factors. To prevent this phenomenon from affecting our experimental results, we decided to bypass the saving and loading steps, but use the tensor data directly in the testing stage. We are intended to test all the defense methods including ours in the same setting to ensure the results are comparable. However, using tensor data in the testing stage is one possible reason of the performance of other previous defense methods worse than reported in the original papers. 

\section{Conclusion}
This paper demonstrates that some defense methods show generalizability when against adversarial attacks, especially the image reconstruction model based on image-to-image translation scheme. Then, we proved that using adversarial samples generated by multiple types of attacks in the training process performs better than only using one specific attack. We then showed that a versatile defense trained using six different adversarial attacks gain an average recovered classification accuracy near to the average of six attack-specific defense models. The result of this paper shows the potential of image-to-image translation based defense methods to be trained into a single versatile defense model that performs stably even when encountering unknown attacks. This type of defense has lower training costs and may have widespread application in real-life setting. The next step of our research is to create an app using this versatile model, and test its usefulness using printed pictures of adversarial samples.
%
% ---- Bibliography ----
%
% BibTeX users should specify bibliography style 'splncs04'.
% References will then be sorted and formatted in the correct style.
%
\bibliographystyle{splncs04}
\bibliography{bibdatabase}
\end{document}